\documentclass[11pt,letterpaper,english]{article}
\usepackage{preamble}

\hypersetup{colorlinks=true,linkcolor=mydarkblue,citecolor=mycitecolor,filecolor=mycitecolor,urlcolor=mydarkblue,pdfview=FitH,
  pdftitle={Learning to Search for Dependencies},
  pdfauthor={},
  }

\usepackage{tikz}
\usetikzlibrary{shapes.gates.logic.US,trees,positioning,arrows,automata,decorations.pathmorphing,chains}

\definecolor{mygreen}{rgb}{0,0.6,0}
\definecolor{mygray}{rgb}{0.5,0.5,0.5}
\definecolor{mymauve}{rgb}{0.58,0,0.82}

\begin{document}
\title{Learning to Search for Dependencies}
\author{Kai-Wei Chang$^1$, He He$^2$, Hal Daum\'e III$^2$, John Langford$^3$\\
$^1$ University of Illinois Urbana-Champaign, IL\\
  {\tt kchang10@illinois.edu} \\ 
$^2$ University of Maryland, College Park, MD \\  
  {\tt \{hhe,hal\}@cs.umd.edu} \\ 
$^3$ John Langford Microsoft Research, New York, NY\\
 {\tt jcl@microsoft.com}
}

\maketitle
\begin{abstract}
We demonstrate that a dependency parser can be built using a credit
assignment compiler which removes the burden of worrying about
low-level machine learning details from the parser implementation.
The result is a simple parser which robustly applies to many languages
that provides similar statistical and computational performance with
best-to-date transition-based parsing approaches, while avoiding
various downsides including randomization, extra feature requirements,
and custom learning algorithms.  
\end{abstract}

\section{Introduction}

Transition-based dependency parsers have a long history, in which many
aspects of their construction have been studied: transition
systems~\cite{nivre03parsing,nivre04parser}, feature
engineering~\cite{koo08semisup}, neural-network
predictors~\cite{danqi14nndep} and the importance of training against
a ``dynamic oracle''~\cite{kuhlmann2011dynamic,goldberg13oracles}.  In
this paper we focus on an understudied aspect of building dependency
parsers: the role of getting the underlying machine learning
technology ``right''.  In contrast to previous approaches which use
heuristic learning strategies, we demonstrate that we can easily build
a highly robust dependency parser with a ``compiler'' that
automatically translates a simple specification of dependency parsing
and labeled data into machine learning updates.

An issue with complex prediction problems is credit assignment: When
something goes wrong do you blame the first, second, or third
prediction?  Existing systems commonly take two strategies:
\begin{compactenum}
\item The system may ignore the possibility that a previous prediction
  may have been wrong. Or ignore that different errors may have different
  costs (consequences). Or that train-time prediction may differ from
  the test-time prediction.  These and other issues lead to
  statistical inconsistency: when features are not rich enough for
  perfect prediction the machine learning may converge suboptimally.
\item The system may use hand crafted credit-assignment heuristics to cope
  with errors the underlying algorithm makes and the long-term outcomes of decisions.  
\end{compactenum}
Here, we show instead that a learning to search compiler
~\cite{daume14imperativesearn} can automatically handle credit
assignment using known
techniques~\cite{daume09searn,ross11dagger,ross14aggrevate,chang15lols}
when applied to dependency parsing.  Dependency parsing is more complex than
previous applications of the compiler and may also be of interest for
other similarly complex NLP problems as it frees designers to worry
about concerns other than low-level machine learning.

The advantage here is the combination of correctness and simplicity
via removal of concerns:
\begin{compactenum}
\item The system automatically employs a cost sensitive learning
  algorithm instead of a multiclass learning algorithm, ensuring the
  model learns to avoid compounding errors.
\item The system automatically ``rolls in'' with the learned policy and ``rolls out'' the dynamic oracle insuring competition with the oracle.
\item Advanced machine learning techniques or optimization strategies are enabled with command-line flags with no additional implementation overhead, such as neural networks or ``fancy'' online learning.
\item The implementation is future-friendly: future compilers may yield a better parser.
\item Train/test asynchrony bugs are removed.  Essentially, you only write the 
	test-time ``decoder'' and the oracle. 
\item The implementation is simple: This one is about $300$ lines of C++ code.
\end{compactenum}

Experiments on standard English Penn Treebank and nine other languages
from CoNLL-X show that the compiled parser is competitive with recent
published results (e.g., an average labeled accuracy of $81.7$ over
$10$ languages, versus $80.3$ for \cite{goldberg13oracles}). 

Altogether, this system provides a strong simple baseline for future
research on dependency parsing, and demonstrates that the compiler
approach to solving complex prediction problems may be of broader interest.

\section{Learning to Search}
\label{sec:l2r}
Learning to search is a family of approaches for solving structured
prediction tasks. This family includes a number of specific algorithms including
the incremental structured perceptron \cite{collins04incremental,huang12structured}, \textsc{Searn} \cite{daume09searn},
\textsc{DAgger} \cite{ross11dagger},
\textsc{Aggrevate} \cite{ross14aggrevate},
and others \cite{daume05laso,xu07beam,xu07planning,ratliff07boosting,syed11reduction,doppa12oss,doppa14hcsearch}.
Learning to search approaches solve structured prediction problems by (1) decomposing the production of the structured
output in terms of an explicit search space (states, actions, etc.);
and (2) learning hypotheses that control a policy that takes actions in
this search space.

\newalgorithm{sequential}%
  {\FUN{RunTagger}(\VAR{words})}%
  {
    \SETST{{output}}{[]}
    \FOR{{$n$} $=\CON{1}$ to \FUN{len}({words})}
      \SETST{ref}{{words}{[}{n}{]}\texttt{.}{true\_label}}
      \SETST{output[n]}{\FUN{predict}({words}{[}{i}{]}, ref, output[:n-1])}
    \ENDFOR
    \STATE \FUN{loss}(\# output[n] $\neq$ words[n]\texttt{.}true\_label)
    \STATE \textbf{return} output
  }


In this work we build on recent theoretical and implementational advances in learning to search
that make development of novel structured prediction frameworks easy and efficient using
``imperative learning to search'' \cite{daume14imperativesearn}.
In this framework, an application developer needs to write (a) a ``decoder'' for the
target structured prediction task (e.g., dependency parsing), (b) an annotation in the decoder that computes
losses on the training data, and (c) a \emph{reference policy} on the training data that returns at any prediction point
a ``suggestion'' as to a good action to take at that state\footnote{Some papers in the past make an implicit or explicit assumption that this reference policy is an oracle policy: for every state, it always chooses the best action (assuming it gets to make all future decisions as well).}.
\begin{figure}[t]
\includegraphics[width=\linewidth]{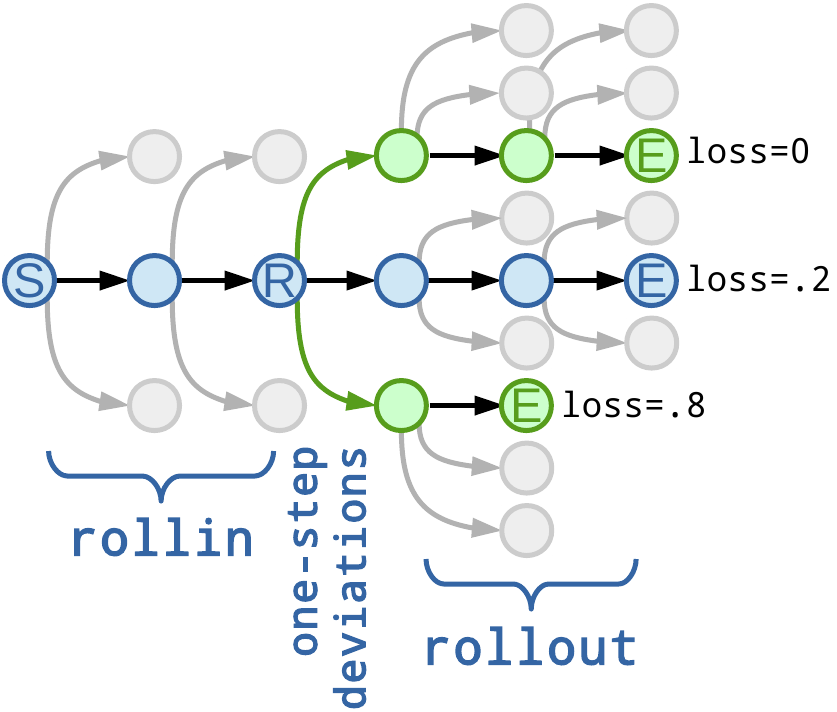}
\caption{A search space implicitly defined by an imperative program. The system begins at the start state $S$ and chooses the middle among three actions by the \textbsf{rollin} policy twice. At state $R$ it considers both the chosen action (middle) and both one-step deviations from that action (top and bottom). Each of these deviations is completed using the \textbsf{rollout} policy until an end state is reached, at which point the loss is collected. Here, we learn that deviating to the top action (instead of middle) at state $R$ decreases the loss by $0.2$.}
\label{fig:searchspace}
\end{figure}

Algorithm~\ref{alg:sequential} shows the code one must write for a part of speech tagger (or generic sequence labeler) under Hamming loss.
The only annotation in this code aside from the calls to the library function \FUN{predict} are the computation of an reference
(an oracle reference is trivial under Hamming loss) and the computation of the total sequence loss at the end of the function.
Note that in this example, the prediction of the tag for the $n$th word depends explicitly on the predictions of \emph{all} previous words!

The machine learning question that arises is how to learn a good \FUN{predict} function given just this information.
The ``imperative learning to search'' answer \cite{daume14imperativesearn} is essentially to run the \FUN{RunTagger} function many times,
``trying out'' different versions of \FUN{predict} in order to learn one that yields low \FUN{loss}. The challenge is how to do this efficiently.
The general strategy is, for some number of epochs, and for each example $(x,y)$ in the training data, to do the following:

\begin{compactenum}
\item Execute \FUN{RunTagger} on $x$ with some \textbsf{rollin policy} to obtain a search trajectory (sequence of action $\vec a$) and loss $\ell_0$
\item Many times:
\begin{compactenum}
\item Choose some time step $t \leq |\vec a|$
\item Choose an alternative action $a_t' \neq a_t$
\item Execute \FUN{RunTagger} on $x$, with \FUN{predict} return $a_{1:t-1}$ initially, then $a_t'$, then acting according to a \textbsf{rollout policy} to obtain a new loss $\ell_{t,a_t'}$
\item Compare the overall losses $\ell_0$ and $\ell_{t,a_t'}$ to construct a classification/regression example that demonstrates how much better or worse $a_t'$ is than $a_t$ in this context
\end{compactenum}
\item Update the learned policy
\end{compactenum}

Figure~\ref{fig:searchspace} shows a schematic of the search space implicitly defined by an imperative program. By executing this program three times (in this example), we are able to explore three different trajectories and compute their losses. These trajectories are defined by the \textbsf{rollin} policy (what determines the initial trajectory), the position of one-step deviations (here, state $R$), and the \textbsf{rollout} policy (which completes the trajectory after a deviation).

By varying the rollin policy, the rollout policy and the manner in which classification/regression examples are created, this general framework can mimic algorithms like \textsc{Searn}, \textsc{DAgger} and \textsc{Aggrevate}. For instance, \textsc{DAgger} uses \textbsf{rollin}=learned policy\footnote{Technically, \textsc{DAgger} rolls in with a mixture which is almost always instantiated to be ``reference'' for the first epoch and ``learned'' for subsequent epochs.} and \textbsf{rollout}=reference, while \textsc{Searn} uses \textbsf{rollin}=\textbsf{rollout}=stochastic mixture of learned and reference policies.

\section{Dependency Parsing by Learning to Search}
\label{sec:dep}
\begin{figure*}
	\resizebox{\textwidth}{!}{
	\begin{tabular}{@{ }l|r@{ }l@{ }l@{}}
		\multirow{2}{*}{Action} & \multicolumn{3}{c}{Configuration}\\
		& {\bf S}tack & {\bf B}uffer & {\bf A}rcs \\
		\hline
		& [{\bf Root}] & [Flying planes can be dangerous]  & \{\}\\
		\textsc{Shift} &[{\bf Root} Flying] & [planes can be dangerous] & \{\}\\		
		\textsc{Reduce-left} & [{\bf Root}] &[planes can be dangerous] & \{(planes, Flying)\}\\
		\textsc{Shift} &[{\bf Root} planes] & [can be dangerous] & \{(planes, Flying)\}\\
		\textsc{Reduce-left} & [{\bf Root}] &[can be dangerous] & \{(planes, Flying), (can, planes)\}\\
		\textsc{Shift} & [{\bf Root} can] &[be dangerous] & \{(planes, Flying), (can, planes)\}\\
		\textsc{Shift} & [{\bf Root} can be] &[dangerous] & \{(planes, Flying), (can, planes)\}\\		
		\textsc{Shift} & [{\bf Root} can be dangerous] &[] & \{(planes, Flying), (can, planes)\}\\		
		\textsc{Reduce-Right} & [{\bf Root} can be] &[] & \{(planes, Flying), (can, planes), (be, dangerous)\}\\		
		\textsc{Reduce-Right} & [{\bf Root} can] &[] & \{(planes, Flying), (can, planes), (be, dangerous), (can, be)\}\\		
		\textsc{Reduce-Right} & [{\bf Root}] &[] & \{(planes, Flying), (can, planes), (be, 
		dangerous), (can, be), ({\bf Root}, can)\}\\		
	\end{tabular}
	}
\\
	\begin{tabular}{cc}		
\resizebox{0.45\textwidth}{!}{
\begin{tikzpicture}[start chain,->,>=stealth',node 
  distance=0.5cm,very thick,every node/.style={anchor=base}]
  \node   (A) at (0,1)   {\bf Root};
  \node   (B) at (1.5,1) {Flying};
  \node   (C) at (3,1) {planes};
  \node   (D) at (4.5,1)  {can};
  \node   (E) at (6,1)  {be};
  \node   (F) at (7.5,1)  {dangerous};
  \path (0,1.5)    edge [bend left=64]   (4.5,1.5)
	    (4.4,1.5) edge [bend right=55]     (3.1,1.5)
	    (2.9,1.5) edge [bend right=55]     (1.6,1.5)		
        (4.6,1.5) edge [bend left=55]     (5.9,1.5)
        (6.1,1.5)   edge [bend left=64]      (7.5,1.5);
\end{tikzpicture}
}&
\resizebox{0.45\textwidth}{!}{
\begin{tikzpicture}[start chain,->,>=stealth',node 
  distance=0.5cm,very thick,every node/.style={anchor=base}]
  \node   (A) at (0,1)   {\bf  Root};
  \node   (B) at (1.5,1) {Flying};
  \node   (C) at (3,1) {planes};
  \node   (D) at (4.5,1)  {can};
  \node   (E) at (6,1)  {be};
  \node   (F) at (7.5,1)  {dangerous};
  \path (0,1.5)    edge [bend left=64]   (4.5,1.5)
	    (1.6,1.5) edge [bend left=55]     (3.1,1.5)
	    (4.4,1.5) edge [bend right=55]     (1.4,1.5)		
        (4.6,1.5) edge [bend left=55]     (5.9,1.5)
        (6.1,1.5)   edge [bend left=64]      (7.5,1.5);
\end{tikzpicture}
}\\
Parse tree derived by the above parser & Gold parse tree
\end{tabular}
	\caption{An illustrative example of an arc-hybrid transition parser.  
	The above table show the actions taken and the intermediate configurations generated by 
	a parser.  The parse tree derived by the parser is in the bottom left, and 
	the gold parse tree is the bottom right.  
	The distance between these two trees is 2.	}

	\label{tab:dep}
\end{figure*}

Learning to search provides a natural framework for implementing a 
transition-based dependency parser. 
A transition-based dependency parser takes a sequence of actions and parses a 
sentence from left to right by maintaining a \emph{stack} $S$, a 
\emph{buffer} $B$, and a set of \emph{dependency arcs} $A$.  
The stack maintains partial parses, the buffer stores the 
words to be parsed, and $A$ keeps the arcs that have been generated so far.  
The configuration of the parser at each stage can be defined by a triple
$(S, B, A)$.
For the ease of notation, we use $w_p$ to represent the leftmost word in the buffer 
and use $s_1$ and $s_2$ to denote the top and the second top 
words in the stack. A dependency arc $(w_h,w_m)$ is a directed edge that indicates word $w_h$
is the parent of word $w_m$. When the parser terminates, the arcs in $A$ form a projective dependency tree.  
We assume that each word only has one parent in the derived dependency parse tree, and use $A[w_m]$ to denote the parent of word $w_m$.
For labeled dependency parsing, we further assign a tag to each arc representing the dependency type between the head and the modifier.  
For simplicity, we assume an unlabeled parser in the following description.
The extension from an unlabeled parser to a labeled parser is straightforward, and is discussed at the end of this section.

\newalgorithm{Trans}%
{\FUN{Trans}($S$, $B$, $A$, action)}%
  {
	\STATE{Let $w_p$ be the leftmost element in $B$}
  	\IF{action = \textsc{Shift}}
  	    \STATE{$S$.push($w_p$)}
  	    \STATE{remove $w_p$ from $B$}
	\ELSIF{action= \textsc{Reduce-Left}}
		\SETST{top}{$S$.pop()}
		\SETST{$A$}{$A \cup$ ($w_p$,top)}
	\ELSIF{action = \textsc{Reduce-Right}}
		\SETST{top}{$S$.pop()}
		\SETST{$A$}{$A \cup$ ($S$.top(), top)}
	\ENDIF
	\STATE{\textbf{return} $S, B, A$}
  }

We consider an arc-hybrid transition 
system~\cite{kuhlmann2011dynamic}\footnote{The learning to search framework is 
also suitable for other transition-based dependency parsing systems, such as 
arc-eager~\cite{nivre03parsing} or arc-standard~\cite{nivre04parser} transition systems.}.
In the initial configuration, the buffer $B$ contains all the words in the sentence,
a dummy root node is pushed in the stack $S$, and the set of arcs $A$ is empty. 
The root node cannot be popped out at anytime during parsing.
The system then takes a sequence of actions until the buffer is empty  and the 
stack contains only the root node (i.e., $|B|=0$ and $S=\{\textbf{Root}\}$). When the process terminates, a parse tree is derived.
At each state, the system can take one of 
the following actions:
\begin{enumerate}
	\item \textsc{Shift}: push $w_p$ to $S$ and move $p$ to the next word. (Valid when $|B|>0$).
\item \textsc{Reduce-left}: add an arc ($w_p$, $s_1$)
	to $A$ and pop $s_1$. (Valid when $|B| > 0$ and $|S| > 1$).
\item \textsc{Reduce-right}: add an arc ($s_2$, $s_1$) 
	to $A$ and pop $s_1$. (Valid when  $|S| > 1$).
\end{enumerate}
Algorithm \ref{alg:Trans} shows the execution of these actions during parsing, and
Figure \ref{tab:dep} demonstrates an example of transition-based dependency parsing.

We can define a search space for dependency parser such that each state represents one configuration 
during the parsing. The start state is associated with the initial 
configuration, and the end states are associated with the configurations that 
$|B|= 0$ and $S=\{\textbf{Root}\}$. The loss of each end state is defined by the 
distance between the derived parse tree and the gold parse tree.  
The above transition actions define how to move from one search state to the 
other.  In the following, we describe our implementation details.

\newalgorithm{depparsing}%
  {\FUN{RunParser}(\VAR{sentence})}%
  {
  \SETST{{stack $S$}}{\{\bf Root\}}
    \SETST{{buffer $B$}}{[words in sentence]}
    \SETST{{arcs $A$}}{$\emptyset$}
	\WHILE{{ $B\neq \emptyset$ or $|S| > 1$}}
	  \SETST{ValidActs}{\FUN{GetValidActions}($S, B$)}
	  \SETST{features}{\FUN{GetFeat}($S, B, A$)}
	  \SETST{ref}{\FUN{GetGoldAction}($S, B$)}
	  \SETST{action}{\FUN{predict}({features, ref, ValidActs})}
	  \SETST{$S, B, A$}{\FUN{Trans}($S, B , A$, action)}
    \ENDWHILE
    \STATE \FUN{loss}($A[w]$ $\neq$ $A^*[w]$, $\forall w \in$ sentence)
    \STATE \textbf{return} output
  }

{\noindent \bf{Implementation}} As mentioned in Section \ref{sec:l2r}, to implement a parser using the learning to search framework, 
we need to provide a decoder, a loss function and reference policy. Thanks to recent work \cite{goldberg13oracles}, we know how to compute a ``dynamic oracle'' reference policy that \emph{is} optimal.
The loss can be measured by how many parents are different between the derived 
parse tree and the gold annotated parse tree.  
Algorithm \ref{alg:depparsing} shows the pseudo-code of a decoder for a {\it unlabeled} dependency parser.
We discuss each subcomponent below.   
\begin{itemize}
\item \textsc{GetValidAction} returns a set of valid actions that can be taken 
based on the current configuration.
\item \textsc{GetFeat} extracts features based on the current configuration.   
The features depend on the top few words in the stack and leftmost few words in the buffer
as well as their associated part-of-speech tags.
We list our feature templates in Table \ref{tab:features}. All features are generated dynamically because configuration changes during parsing.
\item \textsc{GetGoldAction} implements the dynamic oracle described in \cite{goldberg13oracles}. 
The dynamic oracle returns the optimal action in any state that leads to the 
reachable end state with the minimal loss.
\item \textsc{Predict} is a library call implemented in the learning to search system.  
Given training samples, the learning to search system can learn the policy automatically.
Therefore, in the test phase, this function returns the predicted action leading to an end state with small structured loss.
\item \textsc{Trans} function implements the hybrid-arc transition system.  
Based on the predicted action and labels, it updates the parser's configuration, and move the agent to the next search state.
\item \textsc{Loss} function is used to measure the distance between the predicted output and the gold annotation. Here, 
we simply used the number of words for which the parent is wrong as the loss. The \textsc{Loss} has no effect in the test phase.
\end{itemize}

The above decoder implements an unlabeled parser. To build a labeled parser, 
when the transition action is \textsc{Reduce-left} or \textsc{Reduce-right}, 
we call the \textsc{Predict} function again to predict the dependency type of the 
arc. The loss in the labled dependnecy parser can be  measured by $\sum_{w_i} loss(w_i)$, where
\begin{equation}
	\begin{split}
	loss(w_i) = \begin{cases}
		2 &  A[w_i] \neq A^*[w_i] \\
		1 &  A[w_i] = A^*[w_i], L[w_i] \neq L^*[w_i] \\
		0 & \text{Otherwise.}
	\end{cases}
\end{split}
\end{equation}
$A[w_i]$ and $A^*[w_i]$ are the parent of $w_i$ in the derived parse tree and 
gold parse tree, respectively, $L[w_i]$ is the label assign to the arc 
$(A[w_i], w_i)$.  We observe that this simple loss function performs well 
empirically.

\begin{table}
	\begin{tabular}{l}
\hline
 Unigram Features \\ 
\hline
$s_1,s_2, s_3, b_1, b_2, b_3, L_1(s_1), L_2(s_1)$,\\
$R_1(s_1), R_1(s_2),  L_1(b_1), L_2(b_1), L_1(s_2) $ \\ 
\hline
Bigram Features\\
\hline
$s_1  s_1$, $s_2  s_2$, $s_3  s_3$, $b_1  b_1$, $b_2  b_2$, $b_3  b_3$, $s_1  b_1$,\\
$s_1  s_2$, $b_1  b_2$\\
\hline
Trigram Features\\
\hline
$s_1  s_2   s_3$ , $s_1  b_1   b_2$,  $s_1  s_2   b_1$, $s_1  b_1   b_3$,\\
$b_1  b_2   b_3$ ,  $s_1  R_1(s_1)  R_1(s_2)$, $s_1  L_2(s_1)  L_2(b_1)$,\\
$b_1  L_1(b_1)  L_2(b_1)$,  $s_1  s_2   L_1(b_1)$, $s_1  b_1  L_1(s_1)$,\\
$s_1  b_1  L_1(s_2)$,  $s_1  b_1  L_1(b_1)$ \\
	\hline
	\end{tabular}
	\caption{Features used in our dependency parsing system. $s_i$ represents 
	the $i$-th top element in the stack $S$.  $b_i$ is the $i$-th leftmost word in 
	the buffer $B$. $L_i(w)$ and 
	$R_i(w)$ are the $i$-th leftmost child and rightmost child of the word $w$.
	For each feature template, we includes the surface string and the associated 
	part-of-speech (POS) tag as features. For $R_i(w)$ and $L_i(w)$, we also 
	include arc labels as features.  
	A feature hashing technique \cite{weinberger09hashing} is employed to provide a fast feature lookup.  
	}	
	\label{tab:features}
\end{table}

We implemented our parser based on an open-source library supporting learning to 
search.  The implementation requires about 300 lines of C++ code.
The reduction of implementation effort comes from two-folds.
First, in the learning to search framework, there is no need to implement a learning algorithm.
Once the decoding function is defined, the system is able to learn the best ``\textsc{Predict}'' function 
from training data. 
Second, L2S provides a unified framework, which allows the library to serve common functions for ease of implementation.
For example, quadratic and cubic feature generating functions and a feature hashing mechanism are provided by the library. 
The unified framework also allows a user to experiment with different base learners 
and hyper-parameters using command line arguments without modifying the code.

\noindent {\bf Base Learner} As mentioned in Section \ref{sec:l2r},
the learning to search framework reduces structured prediction to
cost-sensitive multi-class classification, which can be further
reduced to regression.  This reduction framework allows us to employ
well-studied binary and multi-class classification methods as the base
learner.  We analyze the value of using more powerful base learners in
the experiment section.

\section{Experimental Results}
While most work compares with MaltParser or MSTParser, which are
indeed weak baselines, we compare with two recent strong baselines:
the greedy transition-based parser with dynamic oracle~\cite{goldberg13oracles}
and the Stanford neural network parser~\cite{danqi14nndep}.
We evaluate on a wide range of different languages,
and show that our parser achieves comparable or better results on all languages,
with significantly less engineering.

\begin{table}[t]
\centering
\begin{tabular}{l|ccc}
\toprule
Parser & Transition & Base learner & Reference \\
\midrule
\our{}& arc-hybrid & NN & Dynamic \\
\dn{}& arc-hybrid & perceptron & Dynamic \\
\nn{}& arc-standard & NN & Static \\
\bottomrule
\end{tabular}
\caption{Parser settings.}
\label{tab:parsers}
\end{table}

\begin{table*}[t!]
\centering
\begin{tabular}{l|cccccccccc|c}
\toprule
Parser & \textsc{Ar} & \textsc{Bu} & \textsc{Ch} & 
\textsc{Da} & \textsc{Du} & \textsc{En} & \textsc{Ja} & \textsc{Po} & 
\textsc{Sl} & \textsc{Sw} & Avg\\
\midrule
& \multicolumn{10}{c}{UAS} & \\
\midrule

\our{} & 77.59 & \textbf{90.64} & \textbf{90.46} & \textbf{88.03} 
& \textbf{78.06} & \textbf{92.30} & 90.89 & \textbf{89.77} & \textbf{81.28} & \textbf{89.12} & \textbf{86.81} \\
\dn{} & \textbf{77.89} & 89.54 & 89.41 & 87.37 & 74.63 & 91.84 & \textbf{92.72} & 85.82 & 77.14 & 87.85 & 85.42 \\
\nn{} & 67.37$^*$ & 88.05 & 87.31 & 82.98 & 75.34 & 90.20 & 89.45 & 83.19$^*$ & 63.60$^*$ & 85.70 & 81.32$^*$ \\
\midrule
& \multicolumn{10}{c}{LAS} & \\
\midrule
\our{} & \textbf{66.44} & \textbf{85.07} & \textbf{86.43} & 81.36 & \textbf{73.55} & \textbf{91.09} & 89.53 & \textbf{84.68} & \textbf{72.48} & \textbf{82.81} & \textbf{81.34} \\
\dn{} & 66.33 & 84.73 & 85.14 & \textbf{82.30} & 70.26 & 90.81 & \textbf{90.91} & 82.00 &  68.65 & 82.21 & 80.33 \\
\nn{} & 51.72$^*$ & 84.01 & 82.72 & 77.44 & 71.96 & 89.10 & 87.37 & 77.88$^*$ & 51.08$^*$ & 80.09 & 75.34$^*$ \\
\bottomrule
\end{tabular}
\caption{UAS and LAS on PTB and CoNLL-X. The average score over all languages is shown in the last column. The best scores for each language is bolded.
\nn{} makes assumptions about the structure of languages and hence obtains substantially worse performance on languages  with multi-root trees (marked with $^*$).   
Excluding these languages, \nn{} achieves 85.6 (UAS) and 81.8 (LAS) in average, while \our{} achieves 88.5 and 84.3.}
\label{tab:result}
\end{table*}

\subsection{Datasets}
We conduct experiments on the English Penn Treebank (PTB)~\cite{Marcus:1993} and 
the CoNLL-X~\cite{buchholz2006conll} datasets for 9 other languages, including
Arabic, Bulgarian, Chinese, Danish, Dutch, Japanese, Portuguese, Slovene and Swedish.
For PTB, we convert the constituency trees to dependencies by the head rules of \newcite{yamada03}.
We follow the standard split: sections 2 to 21 for training,
section 22 for development and section 23 for testing.
The POS tags in the evaluation data is assigned by the Stanford POS tagger~\cite{ToutanovaPos}, which has an accuracy of 97.2\% on the PTB test set.
For CoNLL-X, we use the given train/test splits and reserve the last 10\% of training data for development if needed.
The gold POS tags given in the CoNLL-X datasets are used.

\subsection{Setup and Parameters}
For L2S, the rollin policy is 
a mixture of the current (learned) policy and the reference (dynamic oracle) policy.
The probability of executing the reference policy decreases over each round.
Specifically, we set it to be $1-(1-\alpha)^t$, 
where $t$ is the number of rounds and $\alpha$ is set to $10^{-5}$ in all experiments.
It has been shown~\cite{ross14aggrevate,chang15lols} that when the reference policy is optimal, it is preferable to roll out with the reference.
Therefore, we roll out with the dynamic oracle~\cite{goldberg13oracles}.

Our base learner is a simple neural network with one hidden layer.
The hidden layer size is 5 and we do not use word or POS tag
embeddings.  We find the Follow-the-Regularized-Leader-Proximal (FTRL)
online learning algorithm particularly effective with learning the
neural network and simply use default hyperparameters.

We compare with the recent transition-based parser with dynamic oracles (\dn{})~\cite{goldberg13oracles},
and the Stanford neural network parser (\nn{})~\cite{danqi14nndep}.
Settings of the three parsers are shown in Table~\ref{tab:parsers}.

For \dn{}, we use the software provided by the authors online\footnote{Available at \url{https://bitbucket.org/yoavgo/tacl2013dynamicoracles}}.
Our initial experiments show that its performance is the best using the arc hybrid system 
with exploration parameters $k=1$, $p=1$,
thus we use this setting for all experiments.
The best model evaluated on the development set among 5 runs with different random seeds are chosen for testing.

For \nn{}, we use the latest Stanford parser.\footnote{Available at \url{http://nlp.stanford.edu/software/nndep.shtml}}
Since all other parsers do not use external resources, we do not provide pretrained word embeddings and initialize randomly.
We use the same parameter values as suggested in \cite{danqi14nndep}, 
which are also the default settings of the software.
The best model over 20000 iterations evaluated on the development set is used for testing.\footnote{Enabled by \texttt{-saveIntermediate}.}

In addition, we compare with the RedShift\footnote{Available at
\url{https://github.com/syllog1sm/redshift}} parser on PTB.
For fair comparison, we only use its basic
features (excluding features based on the Brown cluster).
We use the default parameters, which runs a beam search with width 8.
In our experiments, the RedShift parser has UAS 92.10 and LAS 90.83 on the PTB
test set.

\subsection{Results}
We report unlabeled attachment scores (UAS) and labeled attachment scores (LAS) in Table~\ref{tab:result}.
Punctuation is excluded in all evaluations.
Our parser achieves up to 4\% improvement on both UAS and LAS.
Compared with \dn{}, our parser has the same transition system and
oracle but more powerful base learners to choose from. 
Compared with \nn{}, we use much fewer hidden units and
parameters to tune.

 \begin{table}[t]
\centering
\begin{tabular}{l|cc|cc}
\toprule
\multirow{2}{*}{Base Leaner} & \multicolumn{2}{c|}{Dev} & \multicolumn{2}{c}{Test} \\
& UAS & LAS & UAS & LAS \\
\midrule
SGD & 89.34 & 88.03 & 89.34 & 87.89 \\ 
SGD+ & 91.0 & 89.5 & 91.0 & 89.6 \\
NN & 92.02& 90.78 & 91.97 & 90.84\\
NN+FTRL & 92.27& 91.04 & 92.30 & 91.09\\
Multiclass & 91.7& 90.6 & 91.3 & 90.2 \\
\bottomrule
\end{tabular}
\caption{Performance of different base learning algorithms with the L2S parser 
on PTB corpus. 
}
\label{tab:baseLearner}
\end{table}

\noindent{\bf The Value of Strong Base Learners. }
L2S allows us to leverage well studied
classification methods.  
We show the performance when training with base learners using the following 
update rules.  
Unless stated otherwise all the base learners are cost-sensitive multiclass 
classifiers. 
\begin{compactenum}
\item SGD: stochastic gradient descent updates.   
\item SGD+: improved update rule using an adaptive metric~\cite{Adagrad1,Adagrad2},
  importance invariant updates~\cite{Impinvariant}, and normalized
  updates~\cite{Normalized}.
\item NN: a single-hidden-layer neural network with 5 hidden nodes.  
\item NN + FTRL: a neural network learner with follow-the-regularized-leader 
	regularization (the base learner in the above experiments). 
\item Multiclass: a multiclass classifier using NN+FTRL update rules. The gold 
	label is given by the dynamic oracle.   
\end{compactenum}
The results in Table \ref{tab:baseLearner} show that using a strong base learner and taking care of low-level learning details (i.e., using 
cost-sensitive multiclass classifier) can improve the performance.

\begin{table}[t]
\centering
\begin{tabular}{l|cc|cc}
\toprule
\multirow{2}{*}{Base Leaner} & \multicolumn{2}{c|}{Dev} & \multicolumn{2}{c}{Test} \\
& UAS & LAS & UAS & LAS \\
\midrule
Uni-gram & 80.41 & 78.01& 80.97 & 78.65 \\ 
Uni- + Bi-gram &  90.73 &89.46 &91.08 &89.81 \\
All features & 92.27& 91.04 & 92.30 & 91.09\\
\bottomrule
\end{tabular}
\caption{The contribution of Bi-gram and Tri-gram features.
Results are evaluated on the dev and the test set of PTB.  
}
\label{tab:featuresperf}
\end{table}

Finally, Figure \ref{tab:featuresperf} shows the performance of different feature templates.
Using a comprehensive set of features leads to a better dependency parser.

\section{Related Work}
Training a transition-based dependency parser can be viewed as an imitation
learning problem. However, most early works focus on decoding or feature engineering
instead of the core learning algorithm. For a long time, averaged
perceptron is the default learner for dependency parsing.
\newcite{goldberg13oracles} first
proposed dynamic oracles under the framework of imitation learning.
Their approach is essentially a special case of our algorithm:
the base learner is a multi-class perceptron, and no rollout is
executed to assign cost to actions.
In this work, we combine dynamic oracles into learning and explore the
search space in a more
principled way by learning to search:
by cost-sensitive classification, we evaluate the end result of
each non-optimal action instead of treating them as equally bad.

There are a number of works that use the L2S approach to solve
various other structured prediction problems, for example, sequence
labeling~\cite{doppa14hcsearch}, coreference
resolution~\cite{ma14coref}, graph-based dependency
parsing~\cite{he13dep}. However, these works can be considered as a special
setting under our unified learning framework, e.g., with a custom
action set or different rollin/rollout methods. 

To our knowledge, this is the first work that develops a general
programming interface for dependency parsing, or more broadly, for
structured prediction.  Our system bears some resemblance to
probabilistic programming language (e.g., \cite{mccallum09factorie,gordon14probprogram}), however, instead of relying on a
new programming language, ours is implemented in C++ and Python, thus is easily
accessible.

\section{Conclusion and Discussion}
We have described a simple transition-based dependency parser based on
the learning to search framework.  We show that it is now much easier
to implement a high-performance dependency parser.
Furthermore, we provide a wide range of advanced optimization methods
to choose from during training.
Experimental results show that we consistently achieve better
performance across 10 languages.
An interesting direction for future work is to extend the current
system beyond greedy search. In addition, there is a large room for
speeding up training time by smartly choosing where to rollout.




\bibliographystyle{acl2015}
\bibliography{bibfile}
\clearpage
\appendix

\onecolumn

\begin{table}[t]
	\begin{minipage}[t]{0.45\linewidth}
	\begin{tabular}{l|c}
		Function & Number of lines \\
		\hline
		Setup & 90 \\
		\textsc{GetValidActions} & 17\\
		\textsc{GetFeat} & 86 \\
		\textsc{GetGoldAction} & 41\\
		\textsc{Trans} & 28 \\
		\textsc{RunParser} & 40 \\
		\hline
		\textsc{Total} & 331		
	\end{tabular}
	\caption{Number of code lines of our dependency parser implementation.  
 The ``Setup'' contains class 
	constructor, destructor, and handlers for the learning to search framework. }
	\label{tab:loc}
\end{minipage}
\hspace{0.08\linewidth}
	\begin{minipage}[t]{0.45\linewidth}		
	\begin{tabular}{l|c}
		Dependency Parser & Number of lines \\
		\hline
		L2S (ours) & $\sim$300 \\
		Stanford & $\sim$3K \\
		RedShift& \ $\sim$2K \\
		\cite{goldberg13oracles} & $\sim$4K \\
		Malt Parser & $\sim$10K \\
		\\
		\\
	\end{tabular}
	\caption{Number of lines of dependency parser implementations.}
	\label{tab:locp}
\end{minipage}
\end{table}

We implemented our dependency parser at Vowpal Wabbit (\url{http://hunch.  net/~vw/}), a machine learning 
system supporting online learning, hashing, reductions, and L2S. 
Table \ref{tab:loc} shows the number of code lines for each function in our implementation, and Table \ref{tab:locp} shows the number of lines of other popular dependency parsing systems.
Redshift and \newcite{goldberg13oracles} are implemented in Python.   Stanford and 
Malt Parser are in Java. Our implementation is in C++. C++ is usually more lengthy than Python and is 
competitive to Java.   
The code is readable and contains proper comments as shown below.

\begin{lstlisting}
#include "search_dep_parser.h"
#include "gd.h"
#include "cost_sensitive.h"

#define val_namespace 100 // valency and distance feature space
#define offset_const 344429

namespace DepParserTask         {  Search::search_task task = { "dep_parser", run, initialize, finish, setup, nullptr};  }

struct task_data {
  example *ex;
  size_t root_label, num_label;
  v_array<uint32_t> valid_actions, valid_labels, action_loss, gold_heads, gold_tags, stack, heads, tags, temp;
  v_array<uint32_t> children[6]; // [0]:num_left_arcs, [1]:num_right_arcs; [2]: leftmost_arc, [3]: second_leftmost_arc, [4]:rightmost_arc, [5]: second_rightmost_arc
  example * ec_buf[13];
};

namespace DepParserTask {
  using namespace Search;

  void initialize(Search::search& srn, size_t& num_actions, po::variables_map& vm) {
    task_data *data = new task_data();
    data->action_loss.resize(4,true);
    data->ex = NULL;
    srn.set_num_learners(3);
    srn.set_task_data<task_data>(data);
    po::options_description dparser_opts("dependency parser options");
    dparser_opts.add_options()
        ("root_label", po::value<size_t>(&(data->root_label))->default_value(8), "Ensure that there is only one root in each sentence")
        ("num_label", po::value<size_t>(&(data->num_label))->default_value(12), "Number of arc labels");
    srn.add_program_options(vm, dparser_opts);

    for(size_t i=1; i<=data->num_label;i++)
      if(i!=data->root_label)
        data->valid_labels.push_back(i);

    data->ex = alloc_examples(sizeof(polylabel), 1);
    data->ex->indices.push_back(val_namespace);
    for(size_t i=1; i<14; i++)
      data->ex->indices.push_back((unsigned char)i+'A');
    data->ex->indices.push_back(constant_namespace);

	vw& all = srn.get_vw_pointer_unsafe();
    const char* pair[] = {"BC", "BE", "BB", "CC", "DD", "EE", "FF", "GG", "EF", "BH", "BJ", "EL", "dB", "dC", "dD", "dE", "dF", "dG", "dd"};
    const char* triple[] = {"EFG", "BEF", "BCE", "BCD", "BEL", "ELM", "BHI", "BCC", "BJE", "BHE", "BJK", "BEH", "BEN", "BEJ"};
    vector<string> newpairs(pair, pair+19);
    vector<string> newtriples(triple, triple+14);
    all.pairs.swap(newpairs);
    all.triples.swap(newtriples);
    
    srn.set_options(AUTO_CONDITION_FEATURES | NO_CACHING);
    srn.set_label_parser( COST_SENSITIVE::cs_label, [](polylabel&l) -> bool { return l.cs.costs.size() == 0; });
  }

  void finish(Search::search& srn) {
    task_data *data = srn.get_task_data<task_data>();
    data->valid_actions.delete_v();
    data->valid_labels.delete_v();
    data->gold_heads.delete_v();
    data->gold_tags.delete_v();
    data->stack.delete_v();
    data->heads.delete_v();
    data->tags.delete_v();
    data->temp.delete_v();
    data->action_loss.delete_v();
    dealloc_example(COST_SENSITIVE::cs_label.delete_label, *data->ex);
    free(data->ex);
    for (size_t i=0; i<6; i++) data->children[i].delete_v();
    delete data;
  } 

  void inline add_feature(example *ex,  uint32_t idx, unsigned  char ns, size_t mask, uint32_t multiplier){
    feature f = {1.0f, (idx * multiplier) & (uint32_t)mask};
    ex->atomics[(int)ns].push_back(f);
  }

  void inline reset_ex(example *ex){
    ex->num_features = 0;
    ex->total_sum_feat_sq = 0;
    for(unsigned char *ns = ex->indices.begin; ns!=ex->indices.end; ns++){
      ex->sum_feat_sq[(int)*ns] = 0;
      ex->atomics[(int)*ns].erase();
    }
  }

  // arc-hybrid System.
  uint32_t transition_hybrid(Search::search& srn, uint32_t a_id, uint32_t idx, uint32_t t_id) {
    task_data *data = srn.get_task_data<task_data>();
    v_array<uint32_t> &heads=data->heads, &stack=data->stack, &gold_heads=data->gold_heads, &gold_tags=data->gold_tags, &tags = data->tags;
    v_array<uint32_t> *children = data->children;
    switch(a_id) {
      case 1:  //SHIFT
        stack.push_back(idx);
        return idx+1;
      case 2:  //RIGHT
        heads[stack.last()] = stack[stack.size()-2];
        children[5][stack[stack.size()-2]]=children[4][stack[stack.size()-2]];
        children[4][stack[stack.size()-2]]=stack.last();
        children[1][stack[stack.size()-2]]++;
        tags[stack.last()] = t_id;
        srn.loss(gold_heads[stack.last()] != heads[stack.last()]?2:(gold_tags[stack.last()] != t_id)?1:0);
        stack.pop();
        return idx;
      case 3:  //LEFT
        heads[stack.last()] = idx;
        children[3][idx]=children[2][idx];
        children[2][idx]=stack.last();
        children[0][idx]++;
        tags[stack.last()] = t_id;
        srn.loss(gold_heads[stack.last()] != heads[stack.last()]?2:(gold_tags[stack.last()] != t_id)?1:0);
        stack.pop();
        return idx;
    }
    return idx;
  }
  
  void extract_features(Search::search& srn, uint32_t idx,  vector<example*> &ec) {
    vw& all = srn.get_vw_pointer_unsafe();
    task_data *data = srn.get_task_data<task_data>();
    reset_ex(data->ex);
    size_t mask = srn.get_mask();
    uint32_t multiplier = all.wpp << all.reg.stride_shift;
    v_array<uint32_t> &stack = data->stack, &tags = data->tags, *children = data->children, &temp=data->temp;
    example **ec_buf = data->ec_buf;
    example &ex = *(data->ex);

    add_feature(&ex, (uint32_t) constant, constant_namespace, mask, multiplier);
    size_t n = ec.size();

    for(size_t i=0; i<13; i++)
      ec_buf[i] = nullptr;

    // feature based on the top three examples in stack ec_buf[0]: s1, ec_buf[1]: s2, ec_buf[2]: s3
    for(size_t i=0; i<3; i++)
      ec_buf[i] = (stack.size()>i && *(stack.end-(i+1))!=0) ? ec[*(stack.end-(i+1))-1] : 0;

    // features based on examples in string buffer ec_buf[3]: b1, ec_buf[4]: b2, ec_buf[5]: b3
    for(size_t i=3; i<6; i++)
      ec_buf[i] = (idx+(i-3)-1 < n) ? ec[idx+i-3-1] : 0;

    // features based on the leftmost and the rightmost children of the top element stack ec_buf[6]: sl1, ec_buf[7]: sl2, ec_buf[8]: sr1, ec_buf[9]: sr2;
    for(size_t i=6; i<10; i++) {
      if (!stack.empty() && stack.last() != 0&& children[i-4][stack.last()]!=0)
        ec_buf[i] = ec[children[i-4][stack.last()]-1];
    }

    // features based on leftmost children of the top element in bufer ec_buf[10]: bl1, ec_buf[11]: bl2
    for(size_t i=10; i<12; i++)
      ec_buf[i] = (idx <=n && children[i-8][idx]!=0)? ec[children[i-8][idx]-1] : 0;
    ec_buf[12] = (stack.size()>1 && *(stack.end-2)!=0 && children[2][*(stack.end-2)]!=0)? ec[children[2][*(stack.end-2)]-1]:0;

    // unigram features
    uint64_t v0;
    for(size_t i=0; i<13; i++) {
      for (unsigned char* fs = ec[0]->indices.begin; fs != ec[0]->indices.end; fs++) {
        if(*fs == constant_namespace) // ignore constant_namespace
          continue;

        uint32_t additional_offset = (uint32_t)(i*offset_const);
        if(!ec_buf[i]){
          for(size_t k=0; k<ec[0]->atomics[*fs].size(); k++) {
            v0 = affix_constant*((*fs+1)*quadratic_constant + k);
            add_feature(&ex, (uint32_t) v0 + additional_offset, (unsigned char)((i+1)+'A'), mask, multiplier);
          }
        }
        else {
          for(size_t k=0; k<ec_buf[i]->atomics[*fs].size(); k++) {
            v0 = (ec_buf[i]->atomics[*fs][k].weight_index / multiplier);
            add_feature(&ex, (uint32_t) v0 + additional_offset, (unsigned char)((i+1)+'A'), mask, multiplier);
          }
        }
      }
    }

    // Other features
    temp.resize(10,true);
    temp[0] = stack.empty()? 0: (idx >n? 1: 2+min(5, idx - stack.last()));
    temp[1] = stack.empty()? 1: 1+min(5, children[0][stack.last()]);
    temp[2] = stack.empty()? 1: 1+min(5, children[1][stack.last()]);
    temp[3] = idx>n? 1: 1+min(5 , children[0][idx]);
    for(size_t i=4; i<8; i++)
      temp[i] = (!stack.empty() && children[i-2][stack.last()]!=0)?tags[children[i-2][stack.last()]]:15;
    for(size_t i=8; i<10; i++)
      temp[i] = (idx <=n && children[i-6][idx]!=0)? tags[children[i-6][idx]] : 15;	

    size_t additional_offset = val_namespace*offset_const; 
    for(int j=0; j< 10;j++) {
      additional_offset += j* 1023;
      add_feature(&ex, temp[j]+ additional_offset , val_namespace, mask, multiplier);
    }

    size_t count=0;
    for (unsigned char* ns = data->ex->indices.begin; ns != data->ex->indices.end; ns++) {
      data->ex->sum_feat_sq[(int)*ns] = (float) data->ex->atomics[(int)*ns].size();
      count+= data->ex->atomics[(int)*ns].size();
    }
    for (vector<string>::iterator i = all.pairs.begin(); i != all.pairs.end();i++)
      count += data->ex->atomics[(int)(*i)[0]].size()* data->ex->atomics[(int)(*i)[1]].size();	
    for (vector<string>::iterator i = all.triples.begin(); i != all.triples.end();i++)
      count += data->ex->atomics[(int)(*i)[0]].size()*data->ex->atomics[(int)(*i)[1]].size()*data->ex->atomics[(int)(*i)[2]].size();	
    data->ex->num_features = count;
    data->ex->total_sum_feat_sq = (float) count;
  }

  void get_valid_actions(v_array<uint32_t> & valid_action, uint32_t idx, uint32_t n, uint32_t stack_depth, uint32_t state) {
    valid_action.erase();
    if(idx<=n) // SHIFT
      valid_action.push_back(1);
    if(stack_depth >=2) // RIGHT
      valid_action.push_back(2);	
    if(stack_depth >=1 && state!=0 && idx<=n) // LEFT
      valid_action.push_back(3);
  }

  bool is_valid(uint32_t action, v_array<uint32_t> valid_actions) {
    for(size_t i=0; i< valid_actions.size(); i++)
      if(valid_actions[i] == action)
        return true;
    return false;
  }

  size_t get_gold_actions(Search::search &srn, uint32_t idx, uint32_t n){
    task_data *data = srn.get_task_data<task_data>();
    v_array<uint32_t> &action_loss = data->action_loss, &stack = data->stack, &gold_heads=data->gold_heads, &valid_actions=data->valid_actions;

    if (is_valid(1,valid_actions) &&( stack.empty() || gold_heads[idx] == stack.last()))
      return 1;
    
    if (is_valid(3,valid_actions) && gold_heads[stack.last()] == idx)
      return 3;

    for(size_t i = 1; i<= 3; i++)
      action_loss[i] = (is_valid(i,valid_actions))?0:100;

    for(uint32_t i = 0; i<stack.size()-1; i++)
      if(idx <=n && (gold_heads[stack[i]] == idx || gold_heads[idx] == stack[i]))
        action_loss[1] += 1;
    if(stack.size()>0 && gold_heads[stack.last()] == idx)
      action_loss[1] += 1;

    for(uint32_t i = idx+1; i<=n; i++)
      if(gold_heads[i] == stack.last()|| gold_heads[stack.last()] == i)
        action_loss[3] +=1;
    if(stack.size()>0  && idx <=n && gold_heads[idx] == stack.last())
      action_loss[3] +=1;
    if(stack.size()>=2 && gold_heads[stack.last()] == stack[stack.size()-2])
      action_loss[3] += 1;

    if(gold_heads[stack.last()] >=idx)
      action_loss[2] +=1;
    for(uint32_t i = idx; i<=n; i++)
      if(gold_heads[i] == stack.last())
        action_loss[2] +=1;

    // return the best action
    size_t best_action = 1;
    for(size_t i=1; i<=3; i++)
      if(action_loss[i] <= action_loss[best_action])
        best_action= i;
    return best_action;
  }

  void setup(Search::search& srn, vector<example*>& ec) {
    task_data *data = srn.get_task_data<task_data>();
    v_array<uint32_t> &gold_heads=data->gold_heads, &heads=data->heads, &gold_tags=data->gold_tags, &tags=data->tags;
    uint32_t n = (uint32_t) ec.size();
    heads.resize(n+1, true);
    tags.resize(n+1, true);
    gold_heads.erase();
    gold_heads.push_back(0);
    gold_tags.erase();
    gold_tags.push_back(0);
    for (size_t i=0; i<n; i++) {
      v_array<COST_SENSITIVE::wclass>& costs = ec[i]->l.cs.costs;
      uint32_t head = (costs.size() == 0) ? 0 : costs[0].class_index;
      uint32_t tag  = (costs.size() <= 1) ? data->root_label : costs[1].class_index;
      if (tag > data->num_label) {
        cerr << "invalid label " << tag << " which is > num actions=" << data->num_label << endl;
        throw exception();
      }
      gold_heads.push_back(head);
      gold_tags.push_back(tag);
      heads[i+1] = 0;
      tags[i+1] = -1;
    }

	for(size_t i=0; i<6; i++)
      data->children[i].resize(n+1, true);
  }    
  
  void run(Search::search& srn, vector<example*>& ec) {
    task_data *data = srn.get_task_data<task_data>();
    v_array<uint32_t> &stack=data->stack, &gold_heads=data->gold_heads, &valid_actions=data->valid_actions, &heads=data->heads, &gold_tags=data->gold_tags, &tags=data->tags, &valid_labels=data->valid_labels;
    uint32_t n = (uint32_t) ec.size();

    stack.erase();
    stack.push_back((data->root_label==0)?0:1);
    for(size_t i=0; i<6; i++)
      for(size_t j=0; j<n+1; j++)
        data->children[i][j] = 0;
    
    int count=1;
    uint32_t idx = ((data->root_label==0)?1:2);
    while(stack.size()>1 || idx <= n){
      if(srn.predictNeedsExample())
        extract_features(srn, idx, ec);
      get_valid_actions(valid_actions, idx, n, (uint32_t) stack.size(), stack.size()>0?stack.last():0);
      uint32_t gold_action = get_gold_actions(srn, idx, n);

      // Predict the next action {SHIFT, REDUCE_LEFT, REDUCE_RIGHT}
      count = 2*idx + 1;
      uint32_t a_id= Search::predictor(srn, (ptag) count).set_input(*(data->ex)).set_oracle(gold_action).set_allowed(valid_actions).set_condition_range(count-1, srn.get_history_length(), 'p').set_learner_id(0).predict();
      count++;
      
      uint32_t t_id = 0;
      if(a_id ==2 || a_id == 3){
        uint32_t gold_label = gold_tags[stack.last()];
        t_id= Search::predictor(srn, (ptag) count).set_input(*(data->ex)).set_oracle(gold_label).set_allowed(valid_labels).set_condition_range(count-1, srn.get_history_length(), 'p').set_learner_id(a_id-1).predict();
      }
      count++;
      idx = transition_hybrid(srn, a_id, idx, t_id);
    }

    heads[stack.last()] = 0;
    tags[stack.last()] = data->root_label;
    srn.loss((gold_heads[stack.last()] != heads[stack.last()]));
    if (srn.output().good())
      for(size_t i=1; i<=n; i++)
        srn.output() << (heads[i])<<":"<<tags[i] << endl;
  }
}
\end{lstlisting}
\end{document}